\documentclass[letterpaper]{article}
\usepackage{aaai18}
\usepackage{times}
\usepackage{helvet}
\usepackage{courier}
\usepackage{graphicx}
\usepackage{latexsym}
\usepackage{amsmath}
\usepackage{amssymb}
\usepackage[]{mathtools}
\usepackage{booktabs}
\usepackage{multirow}
\usepackage{paralist}
\usepackage[protrusion=true,
            expansion=true,
            final,
            babel, stretch=30, selected=true]{microtype}
\usepackage{comment}
\usepackage{url}
\usepackage{float}
\usepackage[usenames,dvipsnames,table]{xcolor}
\usepackage{siunitx}
\def\sym#1{\ifmmode^{#1}\else\(^{#1}\)\fi}
\restylefloat{table}
\usepackage{cleveref}
\usepackage{etoolbox}
\newtoggle{draft}
\togglefalse{draft}
\iftoggle{draft}{
\newcommand{\ppc}[1]{{\textcolor{BurntOrange}{[#1 \textsc{--pp}]}}}
\newcommand{\mg}[1]{{\textcolor{OliveGreen}{[#1 \textsc{--mg}]}}}
\newcommand{\cc}[1]{{\textcolor{Blue}{[#1 \textsc{--cc}]}}}
\newcommand{\dk}[1]{{\textcolor{Maroon}{[#1 \textsc{--dk}]}}}
}{
\newcommand{\ppc}[1]{ }
\newcommand{\mg}[1]{ }
\newcommand{\cc}[1]{ }
\newcommand{\dk}[1]{ }
}
\renewcommand{\vec}[1]{{\boldsymbol{\mathbf{#1}}}}   %
\DeclareMathOperator*{\argmin}{arg\,min}
\newcommand{\fenda}{\textsc{Fenda}}
\newcommand{\malopa}{\textsc{MaLOPa}}
\newcommand{\disjoint}{\textsc{Disjoint}}
\newcommand{\tied}{\textsc{Tied}}
\newcommand{\add}{\textsc{Add}}
\newcommand{\addmul}{\textsc{AddMul}}
\newcommand{\affine}{\textsc{Affine}}
\newcommand{\rainbow}{\textsc{Rainbow}}

\newcommand{\doctvec}{\textsc{Doc2Vec}}
\newcommand{\numo}[1]{\num[round-mode=places,round-precision=1]{#1}}
\newcommand{\citet}[1]{\citeauthor{#1}~\shortcite{#1}}
\newcommand{\citep}{\cite}

\usepackage{caption}
\usepackage{subcaption}
\usepackage{graphicx}

\title{Actionable Email Intent Modeling with Reparametrized RNNs}

\begin{document}

\author{Chu-Cheng Lin\thanks{This work was done during an internship at Microsoft Research.} \\
  Johns Hopkins University \\
  Baltimore, MD \\
  {\tt clin103@jhu.edu} \\\And
  Dongyeop Kang \\
  Carnegie Mellon University \\
  Pittsburgh, PA \\
  {\tt dongyeok@cs.cmu.edu} \\\And
  Michael Gamon \\
  Microsoft Research \\ 
  Redmond, WA\\
  {\tt mgamon@microsoft.com} \\\AND
  Madian Khabsa\thanks{This work was done when Madian Khabsa was at Microsoft Research.} \\
  Apple Inc \\ 
  Seattle, WA\\
  {\tt madian@apple.com} \\\And
  Ahmed Hassan Awadallah \\
  Microsoft Research \\ 
  Redmond, WA\\
  {\tt hassanam@microsoft.com} \\\And
  Patrick Pantel \\
  Microsoft Research \\ 
  Redmond, WA\\
  {\tt ppantel@microsoft.com}
}
\maketitle
\begin{abstract}
Emails in the workplace are often intentional calls to action for its recipients. We propose to annotate these emails for what action its recipient will take. We argue that our approach of action-based annotation is more scalable and theory-agnostic than traditional speech-act-based email intent annotation, while still carrying important semantic and pragmatic information. We show that our action-based annotation scheme achieves good inter-annotator agreement. 
We also show that we can leverage threaded messages from other domains, which exhibit comparable intents in their conversation, with domain adaptive \rainbow (Recurrently AttentIve Neural Bag-Of-Words). 
On a collection of datasets consisting of IRC, Reddit, and email, our reparametrized RNNs  
outperform common multitask/multidomain approaches
on several speech act related tasks. We also experiment with a minimally supervised scenario of email recipient action classification, and find the reparametrized RNNs learn a useful representation.
\end{abstract}
\section{Introduction}
\label{sec:introduction}
Despite the emergence of many new communication tools in the workplace, email
remains a major, if not the dominant, messaging platform in many corporate settings \cite{agema2015}. Helping people manage and act on their emails can make them more productive.
Recently, Google's system that suggests email replies has gained wide adoption \cite{kannan2016smart}.
We can imagine many other classes of assistance scenarios that can improve worker productivity. For example, consider a system that is capable of predicting your next action when receiving an email. The system could then offer assistance to accomplish that action, for example in the form of a quick reply, adding a task to your to-do list, or helping you take action against another system. 
To build and train such systems, email data sets are essential, but unfortunately public email datasets such as \citet{klimt2004,oard2015} are much smaller than the proprietary data used by Google; and more importantly, they lack any direct information/annotation regarding the recipients' actions.

In this paper, we design an annotation scheme for such actions and have applied it to a corpus of publicly available emails. In order to overcome the data bottleneck for end-to-end training, we leverage other data and annotations that we hypothesize to contain structures similar to email and recipient actions. We apply multitask and multidomain learning, which use domain or task invariant knowledge to improve performance on a specific task/domain \cite{Caruana1997,yang2014}. We show that these secondary domains and tasks in combination with multitask and multidomain learning can help our model discover invariant structures in conversations that improve a classifier on our primary data and task: email recipient action classification.

Previous work in the deep learning literature tackled multidomain/multitask learning by designing an 
encoder that encodes all data and the domain/task description into a \emph{shared} representation space \cite{collobert2008,glorot2011,ammar2016,yang2017}.
The overall model architecture generally is unchanged from the single-domain single-task setting; but the learned representations are now reparametrized to take account of knowledge from additional data and task/domain knowledge. 
In this work, we propose an alternative approach of \emph{model reparametrization}. We train multiple parameter-sharing models across different domains and tasks jointly, without maintaining a shared encoded representation in the network.
We show that reparametrized LSTMs consistently achieve better likelihood and overall accuracy on test data than common domain adaption variants. We also show that the representation extracted from a network instantiated with the shared parameter weights performs well on a previously unseen task. 

The contributions of this paper are: 

First, we designed an annotation scheme for labeling actionable workplace emails, which as we argue in \cref{sec:recipient-actions}, is more amenable to an end-to-end training paradigm, and collected an annotated dataset. 
Second, we propose a family of reparametrized RNNs for both multitask and multidomain learning. 
Finally, we show that such models encode domain-invariant features and, in the absence of sufficient data for end-to-end learning, still provide useful features for scoping tasks 
in an unsupervised learning setting.
\section{Data}
\label{sec:data}
\subsection{The Avocado Dataset}
\label{sec:avocado}
In this study, all email messages we annotate and evaluate on are part of the Avocado dataset \cite{oard2015}, which consists of emails and attachments taken from 279 accounts of a defunct information technology company referred to as ``Avocado''.\footnote{We considered other email corpora such as the Enron corpus \cite{klimt2004}. We decided to use the Avocado dataset because it is the largest and newest one publicly available.} Email threads are reconstructed from the recipients' mailboxes. For the purpose of this paper, we only use complete (thread contains all replies) and linear (every follow-up is a reply to the previous email) threads.\footnote{The summary statistics are in \cref{tbl:stats}.}  %

\subsection{Recipient Actions}
\label{sec:recipient-actions}
Workplace email is known to be highly task-oriented \cite{khoussainov2005,corston2004}.
As opposed to chit chat on the Internet, speaker intents and expected actions on the email are in general very precise. We aim to annotate the actions, which makes our approach differ in a subtle but important way from previous work such as \cite{cohen04}, which is mostly focused on annotating emails for \emph{sender intents}, modeled after illocutionary acts in Speech Act theory \cite{searle1976}. We believe that annotating recipient actions has the following advantages over annotating sender intents: 
First, action based annotation is not tied to a particular speech act taxonomy. The design of such a taxonomy is highly dependent on the system's use cases \cite{traum1999} and definitions of sender intent can be circular \cite{riezler2014}. Even within a single domain such as email, there have been several different sender intent taxonomies \cite{goldstein2006}.  A speech-act-agnostic scheme that focuses on the recipient's action generalizes better across scenarios.
Our annotation scheme also has a lower risk of injected bias because the annotation relies on expected (or even observed) actions performed in response to an email, as opposed to relying on the annotator's intuition about the sender's intent. 
Lastly, while in this paper we rely on annotators for these action annotations, 
many of our annotated actions translate into very specific actions on the computer. Therefore we anticipate intelligent
user interfaces could be used to capture and remind users of such email actions, as in \citet{dredze2008}.

Based on our findings in two pilot runs of email annotations among the authors, we propose the set of recipient actions listed in \cref{tbl:annotation}, which fall in three broad categories:
\begin{description}
\item[Message sending] 
We identify that in many cases, the recipient is most likely to send out another email, either as a reply to the sender or to someone else. As listed in \cref{tbl:annotation}, \textsc{Reply-Yesno}, \textsc{Reply-Ack}, \textsc{Reply-Other}, \textsc{Investigate}, \textsc{Send-New-Email} are actions that send out a new email, either on the same thread or a new one.
\item[Software interaction] In our pilot study we find some of the most likely recipient actions to be interaction with office softwares such as \textsc{Setup-Appointment} and \textsc{Approve-Request}.
\item[Share content] On many occasions, the most likely actions are to share a document, either as an attachment or via other means. We have an umbrella action \textsc{Share-Content} to capture these actions.
\end{description}

\subsection{Data Annotation}
\label{sec:annotate}
\begin{table}[h]
\small
\begin{tabular}{p{.3\linewidth} p{.5\linewidth} }
\toprule
Action & Description  \\
\midrule
\textsc{Reply-Yesno} & Short yes/no reply to a question raised in the previous email \\
\textsc{Reply-Ack} & Simple acknowledgements such as `got it', `thank you.' \\
\textsc{Reply-Other} & Reply to the thread based on information that is available \emph{without doing any additional investigation.} \\
\textsc{Investigate} & Look into some questions/problems to gather the necessary information and reply with that information.   \\
\textsc{Send-New-Email} & Write a new email that is not a reply to the current thread. \\
\textsc{Setup-Appointment} & Set up appointments/cancel appointments.  \\
\textsc{Approve-Request} & Approve requests (typically from subordinates) through an external system such as an expense report system etc. \\
\textsc{Share-Content} & Share content, as an attachment, a link in the email body, or a location on the network that is known to both the sender and recipients \\
\bottomrule
\end{tabular}
\caption{Set of possible recipient actions in our annotation scheme.} 
\label{tbl:annotation}
\end{table}

A subset of the preprocessed email threads described in \cref{sec:avocado} are subsequently annotated. We ask each annotator to imagine that they are a recipient of threaded emails in a workplace environment.
For each message, we ask the annotator to read through the previous messages in the thread, and annotate with the most likely action (in \cref{tbl:annotation}) they may perform if they had been the addressee of that message.
If the most probable action is not defined in our list, we ask the annotators to annotate with an \textsc{Other} action.

A total of $399$ emails from $110$ distinct threads have been annotated by two paid and trained independent annotators. Cohen's Kappa is $0.75$ for the two annotators.
The authors arbitrated the disagreements. 
We include the distribution across the actions in \cref{tbl:annotation}. 

\begin{table*}
\small
\centering
\begin{tabular}{lp{.9\linewidth}}
\toprule
Dataset & Message  \\
\midrule
IRC & \texttt{could somebody explain how i get the oss compatibility drivers to load automatically in ubuntu ?} \\ 
IRC & \texttt{you should try these ones , apt src deb \_\_URL\_\_ unstable/} \\
IRC & \texttt{Ah , cool . Thanks , I 'll try that .} \\
Reddit & \texttt{Does this really appeal to Sanders supporters ? Can one ( or more of you ) explain to me why ? Full disclosure : I do n't pay ATM fees .}  \\
\bottomrule
\end{tabular}
\caption{Some example non-email messages that are likely to elicit actions related to those observed in email data. IRC chats are very task specific. They are mostly about obtaining technical help. Therefore, we observe many conversational turns that start with information requests, followed by delivery of that information. The Reddit dataset, on the other hand, is more diverse: the discussions in \emph{r/politics} more or less pertain to comments on American public policies and politics. We rarely observe messages that require the recipient to take action; but there are requests and deliveries of information which can potentially help learn the underlying representation.}
\label{tbl:cherrypicked}
\end{table*}

\begin{table*}
\small
\centering
\begin{tabular}{lrrrr}
\toprule
Dataset name (type) & \# of threads & \# of messages & Average thread length & Average message length \\
\midrule
Avocado (Email) & \num{50890} & \num{121917} & \numo{2.3957} & \numo{72.969142941509389} \\
r/politics (Reddit) & \num{15813} & \num{42952} & \numo{2.7162461266046924} & \numo{31.421726578506238}  \\
Ubuntu Dialog (IRC) & \num{50812} & \num{416721} & \numo{8.2012}  & \numo{12.740370655666501} \\
\bottomrule
\end{tabular}
\caption{Statistics of conversational data used in this paper. During preprocessing we truncate each message to \num{256} words, including \textsc{bos} and \textsc{eos} symbols; and each thread to \num{32} messages. The original Ubuntu dataset is much larger (with $\approx$ \num{500000} threads). We truncated it to match the Avocado dataset size for faster training and evaluation of our model. }
\label{tbl:stats}
\end{table*}

\subsection{Additional Domains}
\label{sec:outdomain}

The annotations we collect are comparable in size to other speech act based annotation datasets.
However like other expert-annotated datasets, ours is not large enough for end-to-end training. Therefore, we aim to enrich our training with additional semantic and pragmatic information derived from other tasks and domains without annotation for expected action. We consider data from the following additional domains for multidomain learning: 
\begin{description}
\item[IRC] The Ubuntu Dialog Corpus is a curated collection of chat logs from Ubuntu's Internet Relay Chat technical support channels \cite{lowe2015}.
\item[Reddit] Reddit is an internet discussion community consisting of several \emph{subreddits}, each of which is more or less a discussion forum pertaining to a certain topic. We curate a dataset from the subreddit  \emph{r/politics} over two consecutive months. Each entry in our dataset consists of the post title, an optional post body, and an accompanying tree of comments. We collect linear threads by recursively sampling from the trees.
\end{description}
Messages from IRC and Reddit are less precise in terms of speaker intents; and our recipient action scheme is not directly applicable to them. However, previous studies on speech acts in Internet forums and chatrooms have shown that there are speech acts common to all these heterogeneous domains, e.g. information requests and deliveries. Some such examples are listed in \cref{tbl:cherrypicked}. \cite{arguello2015,moldovan2011} 
We hypothesize that more data from these domains will help recognition of these speech acts, which in turn help recognize the resulting recipient actions.

In all experiments in \cref{sec:experiments}, we use  half of the  dataset as training data, a quarter as the validation data and the remaining quarter as test data. 

\subsection{Metadata-Derived Prediction Tasks}
\label{sec:metadata}
The datasets introduced in \cref{sec:avocado,sec:outdomain} are largely unlabeled as far as recipient actions are concerned, except for the small subset of Avocado data that was manually annotated. However we can still extract useful information from their metadata, such as inferred end-of-thread markers or system-logged events
that can help us formulate additional prediction tasks for a multitask learning setting (listed in \cref{tbl:featuredesc}). We also use these multitask labels to evaluate our multitask/domain model in \cref{sec:end-to-end}.
\begin{table}
\small
\centering
\begin{tabular}{ccp{.6\linewidth}}
\toprule
Identifier & Dataset & Description \\
\midrule
\textsc{e-t} & Email & end of an email thread \\
\textsc{e-a} & Email & this message has attachment(s) \\
\textsc{i-t} & IRC & turntaking \\
\textsc{r-t} & Reddit & end of a Reddit thread \\
\bottomrule
\end{tabular}
\caption{Description of additional prediction labels for multitask learning that we extracted from datasets introduced in {\protect\cref{sec:data}}.}
\label{tbl:featuredesc}
\end{table}

\section{Modeling Threaded Messages}
\subsection{Notations}
\label{sec:notation}
We model threaded messages as a two-layer hierarchy: at the lower layer we have a \emph{message} $\vec{m}$ consisting of a list of words: $\vec{m} = [w_{1\ldots |\vec{m}|}]$. And in turn, a \emph{thread} $\vec{x}$ is a list of messages: $\vec{x} = [\vec{m}_{1\ldots |\vec{x}|}]  \in \mathcal{X}$.
We assume each message thread to come from a specific domain; and therefore define a many-to-one mapping $f(\vec{x})=d$ where $d \in \mathcal{D}$ is the set of all domains.
We also define the tasks $t\in \mathcal{T}$ to have a many-to-one mapping $g(t) = d, d \in \mathcal{D}$. %
For prediction we define the \emph{predictor} of task $t$ as $h_t(\vec{x}): \mathcal{X} \rightarrow \mathcal{Y}$, which predicts sequential tags $\vec{y} = [ y_1 \ldots y_{|\vec{x}|} ] \in \mathcal{Y}$ from a thread $\vec{x}$ on (a valid) task $t$. We also define the real-valued \emph{task loss} of task $t$ on thread $\vec{x}$ to be $\ell_t(\vec{y}, \hat{\vec{y}}): \mathcal{Y} \times \mathcal{Y} \rightarrow \mathbb{R}$, where $\hat{\vec{y}} \in \mathcal{Y}$ is the ground truth.

\subsection{Definition of Multitask/domain Loss}
In this paper, we define the \emph{multitask loss} $l_{\textsc{mt}}$ as the sum of task losses of tasks $\mathcal{T}_d$ under the same domain $d$ for a single (output, ground truth) pair $(\vec{y}, \hat{\vec{y}})$:
\begin{align*}
l_{\textsc{mt}}(\mathcal{T}_d, \vec{y}, \hat{\vec{y}} ) = \sum_{t \in \mathcal{T}_d} \ell_t(\vec{y}, \hat{\vec{y}}),
\end{align*}
and the aggregate loss 
\begin{align*}L_{\textsc{mt}}(  \mathcal{T}_d, \{  \vec{y}^{(d)}_{1 \ldots K_d}, \hat{\vec{y}}^{(d)}_{1 \ldots K_d}   \}) = \sum_{k=1}^{K_d} l_{\textsc{mt}}( \mathcal{T}_d, \vec{y}^{(d)}_k, \hat{\vec{y}}^{(d)}_k )\end{align*} 
is the sum over $K_d$ examples $\vec{y}^{(d)}_{1} \ldots \vec{y}^{(d)}_{K_d}$.

We also define the \emph{multidomain loss} $L_{\textsc{md}}$ to be the sum of aggregate losses over $\mathcal{D}$:
\begin{align}
L_{\textsc{md}}(\{\{  \vec{y}^{(d)}_{1 \ldots K_d}, \hat{\vec{y}}^{(d)}_{1 \ldots K_d}   \} \mid d \in \mathcal{D} \} ) &= \nonumber \\
\sum_{d \in D} L_{\textsc{mt}} (\mathcal{T}_d, \{  \vec{y}^{(d)}_{1 \ldots K_d}, \hat{\vec{y}}^{(d)}_{1 \ldots K_d}   \})
\label{eqn:mdmt}
\end{align}

\subsection{The Recurrent AttentIve Neural Bag-Of-Words model (\rainbow)}
\label{sec:r-nbow}
We start with the {\bf R}ecurrent {\bf A}ttent{\bf I}ve {\bf N}eural {\bf B}ag-{\bf O}f-{\bf W}ord model (\rainbow) as the baseline model of threaded messages. From a high level view, \rainbow\,is a hierarchical neural network with two encoder layers: the lower level encoder is a neural bag-of-words encoder that encodes each message $\vec{m}$ 
into its message embeddings $b(\vec{m})$. And in turn, the upper level encoder transforms the independently encoded message embeddings $[b(\vec{m}_1)\ldots b(\vec{m}_{|\vec{x}|})]$ into thread embeddings via a learned recurrent neural network $\vec{e}_1\ldots \vec{e}_{|\vec{x}|} = f_{\mathrm{RNN}}(b(\vec{m}_1)\ldots b(\vec{m}_{|\vec{x}|}))$.\footnote{There is a slight abuse of annotation since $f_{\mathrm{RNN}}$ actually differs for $\vec{x}$ of different lengths.} \rainbow\,has three main components: message encoder, thread encoder, and predictor.

\paragraph{Message encoder.} We implement the message encoder $b(\vec{m})$ as a bag of words model over the words in $\vec{m}$. Motivated by the unigram features in previous work on email intent modeling, we also add an attentive pooling layer \cite{rush2015} to pick up important keywords. The averaged embeddings then undergo a nonlinear transformation:
\begin{equation}
\mathrm{b}(\vec{m}) = q\left( \sum_{w \in \vec{m}} \frac{\exp ( a(\vec{emb}(w) )) }{\sum_{w' \in \vec{m}} \exp ( a(\vec{emb}(w')))} \vec{emb}(w)\right),
\label{eq:bow}
\end{equation}%
where $q: \mathbb{R}^{n} \rightarrow \mathbb{R}^{h}$ is a learned feedforward network, $\vec{emb}: \mathbb{N} \rightarrow \mathbb{R}^{n}$ is the word embeddings of $w$ and $a: \mathbb{R}^{n} \rightarrow \mathbb{R}$ is the (learned) attentive network that judges how much each word $w$ contributes towards the final representation $b(\vec{m})$.\footnote{There may be concerns about the unordered nature of the neural bag-of-words (NBOW) model. However it has been shown that with a deep enough network, an NBOW model is competitive against syntax-aware RNN models such as Tree LSTMs\cite{tai2015}. In preliminary experiments we did not find the difference between an NBOW and an RNN to be substantial. But the NBOW architecture trains much faster.
}
\paragraph{Thread encoder and predictor.}
\label{sec:output-layers}
The message embeddings are passed onto the thread-level LSTM to produce a \emph{thread embeddings} vector:
\begin{align*}
[\vec{e}_1 \ldots \vec{e}_{|\vec{x}|}] = 
r(b(\vec{m}_1) \ldots b(\vec{m}_{|\vec{x}|}))
\end{align*}

Thread embeddings are then passed to the predictor layer. In this paper, the predictions are distributions over possible labels. We therefore define the predictor $h_t$ to be a $2$-layer feed forward network $p$ that maps thread embeddings to distributions over $V_t$, the label set of task $t$: $h_t(\vec{e}_1 \ldots \vec{e}_{|\vec{x}|}) = [ p(\cdot \mid \vec{e}_1) \ldots p(\cdot \mid \vec{e}_{|\vec{x}|}) ]$.
The accompanying loss is naturally defined as the cross entropy between the predictions $p(\vec{e}_1) \ldots p(\vec{e}_{|\vec{x}|})$ and the empirical distribution $\hat{\vec{y}} = \hat{y}_{1 \ldots |\vec{x}|}$:
\begin{align}
\ell_{t}(\hat{\vec{y}}, \vec{e}_{1\ldots |\vec{x}|}) = \frac{\sum_{i=1}^{|\vec{x}|} - \hat{y}_i \log p(\hat{y}_i \mid \vec{e}_i)}{|\vec{x}|}.
\label{eqn:ace}
\end{align}

\subsection{Multi-Task RNN Reparametrization}
\label{sec:reparametrization}
\begin{figure*}[h]
\centering
\begin{subfigure}[b]{.45\linewidth}
\includegraphics[height=16em]{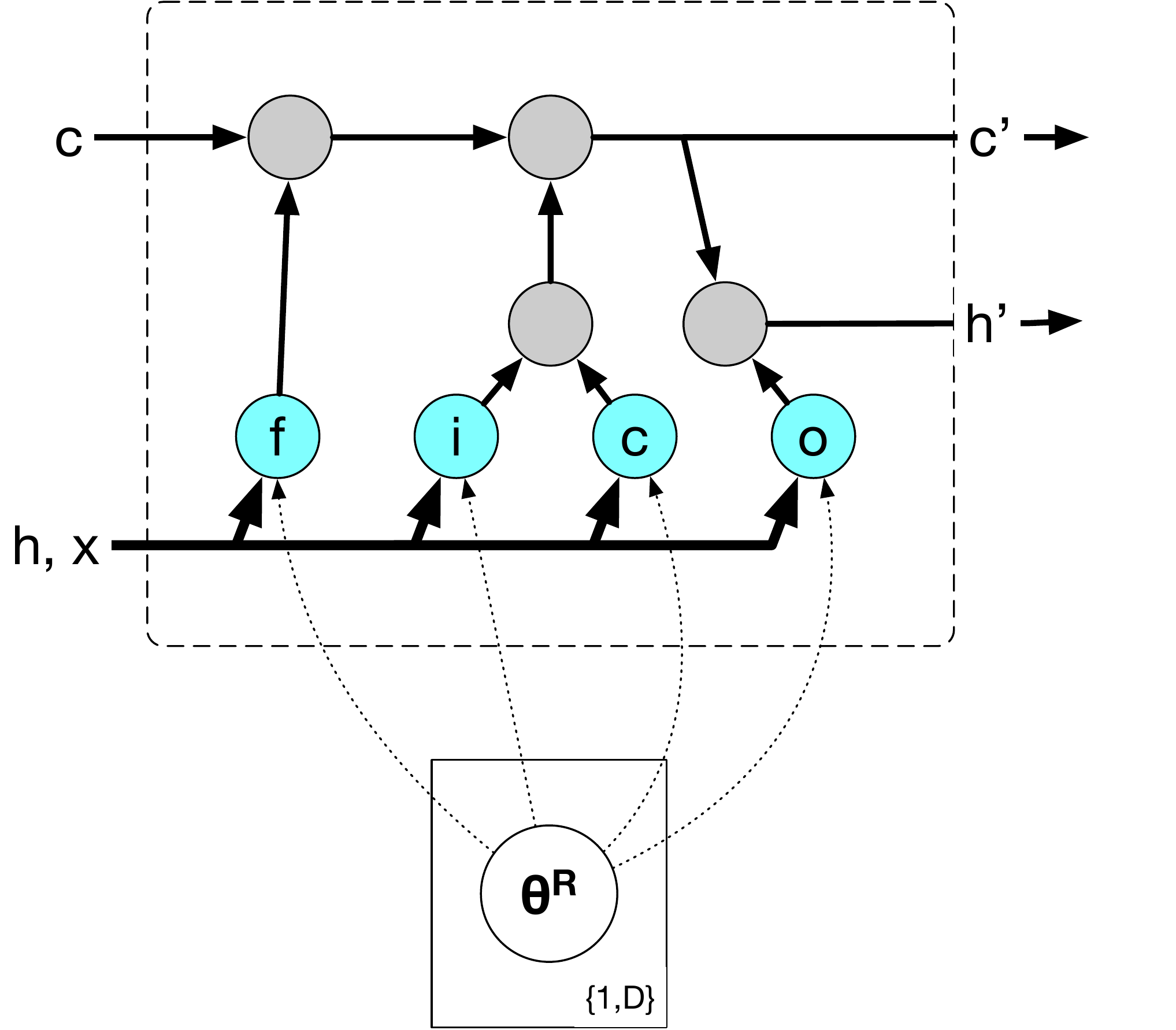}
\caption{LSTM cell}
\label{fig:lstm-vanilla}
\end{subfigure}
\begin{subfigure}[b]{.45\linewidth}
\includegraphics[height=16em]{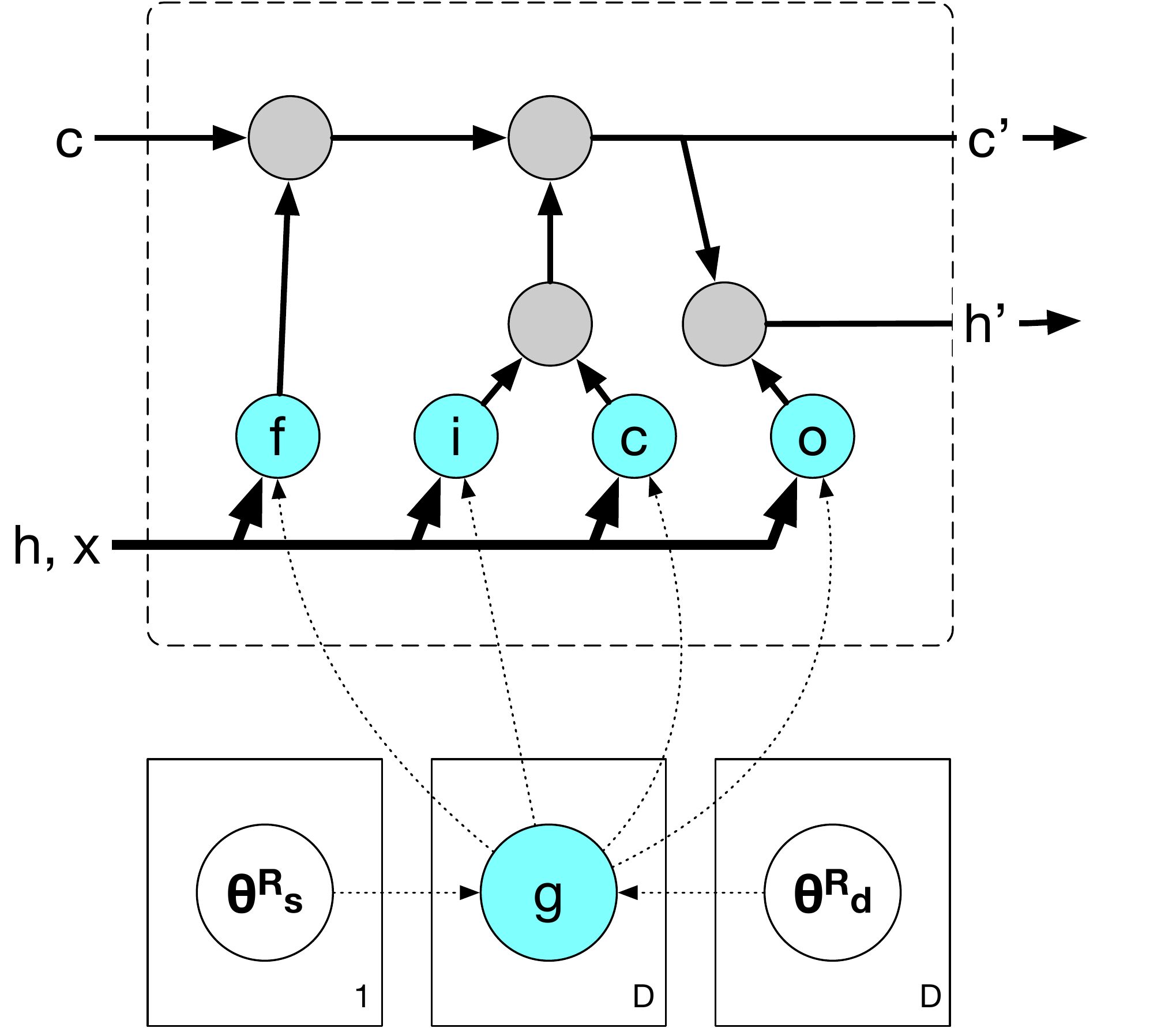}
\caption{Parameter-sharing LSTM cell}
\label{fig:lstm-shared}
\end{subfigure}
\caption{A comparison between partial computation graphs of a single (vanilla) LSTM cell, and our proposed parameter-sharing variants described in \cref{sec:reparametrization}. White circles are learned parameters. Dotted connections indicate parametrization. Parametrized and non-parametrized functions are indicated with blue and gray circles respectively. To model sequences from multiple domains, the conventional LSTM (depicted in \cref{fig:lstm-vanilla}) either shares everything with a set of parameters (the \tied\, setup; $i=1$) or do not share parameters at all (the \disjoint\, setup; $i=D$). In contrast, our parameter-sharing variant in \cref{fig:lstm-shared} models domain-invariant parameters with $\vec{\theta}_s$ and domain-specific parameters with $\left\{ \vec{\theta}_d \right\}$. }\label{fig:parameter-sharing-diagram}
\end{figure*}
\textsc{Rainbow} is an 
extension of Deep Averaging Networks \cite{iyyer2015} to threaded message modeling. It works well for tagging threaded messages for the messages' properties, such as conversation-turn marking in online chats and end-of-thread detection in emails. However, in its current form, the model is trained to work on exactly one task. It also does not capture the shared dynamics of these different domains \emph{jointly} when given out-of-domain data. In this section we describe a family of reparametrized recurrent neural networks that easily accommodates %
multi-domain multi-task learning settings. 

In general, recurrent neural networks take a sequence of 
input data $\vec{x}$ and recurrently apply a nonlinear function, to get
a sequence of transformed representation $\vec{h}$. Here we denote such
transformation with the function $f_{\mathrm{RNN}}$ parametrized by  the RNN parameters $\vec{\theta}^{R}$ as $
\vec{h} = f_{\mathrm{RNN}}(\vec{x}; \vec{\theta}^{R})$.
For an LSTM model, $\vec{\theta}^{R}$ can be formulated as the concatenated vector of input, output, forget and cell gate parameters $[\vec{W}_i, \vec{W}_o, \vec{W}_f, \vec{W}_c]$. And in general, the goal of training an RNN is to find the optimal real-valued vector $\hat{\vec{\theta}}^{R}$ such that $\hat{\vec{\theta}}^{R} = \argmin_{\vec{\theta}^{R}} L(f_{\mathrm{RNN}}(\vec{x}; \vec{\theta}^{R}))$, for a given loss function $L$.

In the context of multidomain learning, we parametrize \cref{eqn:mdmt} in a similar fashion:
\begin{align*}
L_{\textsc{md}}(\{ \{  \vec{y}^{(d)}_{1 \ldots K_d}, \hat{\vec{y}}^{(d)}_{1 \ldots K_d} \} \mid d \in \mathcal{D} \}) = \\
L_{\textsc{md}}(\{ \{  h(\vec{x}_1^{(d)})\ldots h(\vec{x}_{K_d}^{(d)}), \hat{\vec{y}}^{(d)}_{1 \ldots K_d} \} \mid d \in \mathcal{D} \}) =  \\ \sum_{t, \vec{x}, \hat{\vec{y}}} \ell_t(h(\vec{x}), \hat{\vec{y}}; \vec{\theta}^{R}_{t}) &.
\end{align*}
Here we are faced with two modeling choices (depicted in \cref{fig:lstm-vanilla}): we can either model every task $t$ \disjoint ly or with \tied\,parameters.
The \disjoint\, approach learns a separate set of parameters $\vec{\theta}^{R}_t$ per task $t$. Therefore, performance of a task is little affected by data from other domain/tasks, except for the regularizing effect through the word embeddings.

On the other hand the \tied\, approach ties parameters of all domains to a single $\vec{\theta}^{R}$, which
has been a popular choice for multitask/domain modeling --- it has been found that the RNN 
often learns to encode a good shared representation when trained jointly for different tasks \cite{collobert2011,yang2016}. The network also seems to generalize over different domains, too \cite{ragni2016,peng2016}. However it hinges on the assumption that either all domains are similar, or the network is capable enough to capture the dynamics of data from all domains at the same time.

In this paper we propose an alternative approach. Instead of having a single set of parameters $\hat{\vec{\theta}}^{R}$ for all domains, we propose to reparametrize $\vec{\theta}^{R}$ as a function $\phi$ of {\bf s}hared components $\vec{\theta}^{R_s}$ and {\bf d}omain specific components $\vec{\theta}^{R_d}$. Namely:
\begin{equation}
\vec{\theta}^{R} = \phi(\vec{\theta}^{R_s}, \vec{\theta}^{R_d}),
\end{equation}
and our goal becomes minimizing the loss w.r.t both $\vec{\theta}^{R_s}, \vec{\theta}^{R_d}$:
\begin{equation}
\hat{\vec{\theta}}^{R_s}, \hat{\vec{\theta}}^{R_d} = \argmin_{\vec{\theta}^{R_s}, \vec{\theta}^{R_d}} \sum \ell_t(\vec{x}, \hat{\vec{y}}; \vec{\theta}^{R_s}, \vec{\theta}^{R_d}).
\end{equation}
A comparison between the vanilla RNN and our proposed modification can be found in \cref{fig:parameter-sharing-diagram}.
This reparametrization allows us to share parameters among networks trained on data of different domains with the shared component $\vec{\theta}_s$, while allowing the network to work differently on data from each domain with the domain specific parameters $\vec{\theta}_d$.

The design of the function $\phi$ requires striking a balance between model flexibility and generalizability. In this paper we consider the following variants of $\phi$:

\paragraph{Additive (\add)} First we consider $\phi$ to be a linear interpolation of a shared base $\vec{\theta}^{R_s}$ and a network specific component $\vec{\theta}^{R_d}$: 
\begin{equation}
\vec{\theta}^{R} =  \phi_\add (\vec{\theta}^{R_s}, \vec{\theta}^{R_d}; u_d) = \vec{\theta}^{R_s} + \exp(u_d) \vec{\theta}^{R_d},\end{equation}
where $u_d \in \mathbb{R}$. In this formulation \add\,we learn a shared $\vec{\theta}^{R_s}$, and additive domain-specific parameters $\{\vec{\theta}^{R_d} \mid d \in \mathcal{D} \}$ for each domain. We also learn $u_d$ for each domain $d \in \mathcal{D}$, which controls how much effect $\vec{\theta}^{R_d}$ has on the final parameters.

Both \disjoint\, and \tied\,can be seen as degenerate cases of \add:  we recover \disjoint\,when the shared component is a zero vector: $\vec{\theta}^{R_s} = \vec{0}$ %
And with $u_d \rightarrow - \infty$ we have $\vec{\theta}^{R} = \vec{\theta}^{R_s}$, namely \tied. %

\paragraph{Additive + Multiplicative (\addmul)} \add\, has no nonlinear interaction between $\vec{\theta}^{R_s}$ and $\vec{\theta}^{R_d}$: they have independent effects on the composite $\vec{\theta}^{R}$. In \addmul\, we have two components in $\vec{\theta}^{R_d} = [\vec{\theta}^{R_{da}}, \vec{\theta}^{R_{dm}}]$: the additive component $\vec{\theta}^{R_{da}}$ and the multiplicative component $\vec{\theta}^{R_{dm}}$ which introduces nonlinearity without significantly increasing the parameter count: %
\begin{align}
\vec{\theta}^{R} = \phi_{\addmul}(\vec{\theta}^{R_s}, \vec{\theta}^{R_d}; u_d, v_d) & \nonumber \\
= \vec{\theta}^{R_s} + \exp(u_d)\vec{\theta}^{R_{da}} + \exp(v_d)\vec{\theta}^{R_{dm}} \otimes \vec{\theta}_s,\end{align}
where $\otimes$ is the Hadamard product and $u_d,v_d \in \mathbb{R}$ are learned parameters as in the \add\,formulation. 

\paragraph{Affine (\affine)} In this formulation $\vec{\theta}^{R_d}$ are seen as \emph{task embeddings}. We apply a learned affine transformation $\vec{W}$ to the task embeddings and add up the shared component $\vec{\theta}^{R_s}$:
\begin{equation}
\vec{\theta}^{R} = \affine(\vec{\theta}^{R_s}, \vec{\theta}^{R_d}; \vec{W}) = \vec{\theta}^{R_s} + \vec{W} \vec{\theta}^{R_d},  \end{equation}
where $\vec{W}$ is a learned parameter.

\subsection{Optimization}
\label{sec:optimization}
We optimize for the multidomain loss as defined in \cref{eqn:mdmt} with gradient descent methods.
To update parameters, we sample one thread from each domain $\{\vec{m}_{d} \mid d \in \mathcal{D} \}$ and optimize the network parameters with the ADAM optimizer.\cite{kingma2014}
\section{Experiments}
\label{sec:experiments}
\subsection{Evaluation Metrics}
\label{sec:metrics}
In this section we evaluate \rainbow\,and
its multitask/multidomain variants on the datasets we introduced in \cref{sec:data}. We also apply our extracted thread embeddings on a real-world task setting of email action classification with impoverished resources.

Probabilistic models are usually evaluated on the log-likelihood of the test data $S=\{(\vec{x}_1, \hat{\vec{y}}_1) \ldots (\vec{x}_{|S|}, \hat{\vec{y}}_{|S|}) \}$: $\sum_{(\vec{x}, \hat{\vec{y}}) \in S} \log p(\hat{\vec{y}} \mid \vec{x}) $. However, in our multidomain setting
we have multiple datasets that differ in size and average sequence length. Therefore we evaluate our models on mean average cross entropy (MACE):
\begin{align}
\MoveEqLeft \mathrm{MACE}(S) = \nonumber \\ &\frac{ - \sum_{(\vec{x}, \hat{\vec{y}}) \in S} (1/|\vec{y}|) \cdot \left( \sum_{i=1}^{|\hat{\vec{y}}|} \log p_t(\hat{y}_i \mid \vec{e}) \right) }{|S|},
\end{align}
where $\vec{e}$ are the thread embeddings of $\vec{x}$, and  $p_t(\cdot \mid \vec{e})$ follow the definition in \cref{sec:r-nbow}. MACE normalizes by both sequence length $|\vec{y}|$ and dataset length $|S|$: a model that ignores the resource-poor tasks or short sequences tends to perform poorly under this metric. MACE can therefore be seen as per-task (log) perplexity: a larger MACE value means the model performs worse on the dataset; and the oracle would obtain a MACE value of $0$. The average of MACE scores also has the natural interpretation of the geometric mean of log likelihoods over different tasks/domains. In addition to MACE, we also evaluate on accuracy in \cref{tbl:mdmt-average}.

All experiments in \cref{sec:experiments} are trained on train splits. For experiments in \cref{sec:ablation,sec:end-to-end} we evaluated on metadata-derived features in \cref{tbl:featuredesc}.
After each epoch of training, the model is evaluated on the validation split to check if the performance has stopped increasing. The training procedure terminates when no new performance gains are observed for two consecutive epochs.

\subsection{Effectiveness of \rainbow: Ablation Studies}
\label{sec:ablation}
We evaluate \rainbow~by comparing it, in the single task setup, against two simpler variant architectures: one is taking away the {\bf r}ecurrent thread encoder ({\bf -R}), the other is replacing the {\bf a}ttentive pooling layer with an unweighted mean ({\bf -A}). We evaluate the four configurations on the four labels listed in \cref{tbl:featuredesc} and report the averaged MACE numbers in \cref{tbl:ablation}. We find that both attentive pooling and the recurrent network help; but the latter has a much more pronounced effect.
\rainbow~without the two additions ({\bf -R}, {\bf -A}) is reduced to the vanilla Deep Average Network model, a neural baseline that has been shown to be competitive against other neural and non-neural models. 

\begin{table}[h]
\small
\centering
\begin{tabular}{crr}
\toprule
Configuration  & {\bf +R} & {\bf -R} \\
\midrule
{\bf +A} & \num{0.079575433335245443} & \num{0.11627699018121548} \\
{\bf -A} & \num{0.080006082138796231} & \num{0.11739763048763892} \\
\bottomrule
\end{tabular}
\caption{MACE values of the \rainbow ablation tests (lower is better). {\bf +/-R} and {\bf +/-A} indicates the presence/absence of the thread encoder and the attentive pooling layer, respectively.
}
\label{tbl:ablation}
\end{table}

\subsection{Multidomain/task Experiments}
\label{sec:end-to-end}
\begin{table*}
\small
\centering
\begin{tabular}{crrrrrrrrrr}
\toprule
Task  & \multicolumn{2}{c}{E-T} & 
\multicolumn{2}{c}{E-A} &
\multicolumn{2}{c}{I-T} & 
\multicolumn{2}{c}{R-T} & \multicolumn{2}{c}{Average} \\
\midrule
 & MACE & Acc & MACE & Acc & MACE & Acc & MACE & Acc & MACE & Acc \\
\midrule
\multicolumn{11}{c}{Aggregated Results} \\
\midrule
\add & {\bf \num{0.0918734521479}}  & {\bf \numo{80.970119836}} & \num{0.0508520863839}  & \numo{93.3558025859} & {\bf \num{0.0741377006254}}  & \numo{22.3333070928} & {\bf \num{0.103938890954}}  & {\bf \numo{67.9553561712}}  & {\bf \num{0.320802130111}} & \numo{66.1536464215}   \\ 
\addmul & \num{0.0926710116765}  & {\bf \numo{80.970119836}} & \num{0.0509775824498}  & \numo{93.3183538316} & \num{0.0742387450691}  & \numo{22.4461413314} & \num{0.105045574837}  & \numo{67.6426865555}  & \num{0.322932914033} & \numo{66.0943253886}   \\ 
\affine & \num{0.0929663368405}  & \numo{80.9563229265} & {\bf \num{0.0502171702779}}  & \numo{93.3755124566} & \num{0.074114068015}  & \numo{22.682306017} & \num{0.105462913442}  & \numo{66.1934347406}  & \num{0.322760488575} & \numo{65.8018940352}   \\ 
\midrule
\disjoint & \num{0.0933029980277}  & \numo{80.9464679912} & \num{0.0507003854061}  & {\bf \numo{93.3932513403}} & \num{0.0741911718307}  & \numo{22.3752919258} & \num{0.107783046982}  & \numo{65.4267484813}  & \num{0.325977602247} & \numo{65.5354399346}   \\ 
\tied & \num{0.0936650097902}  & \numo{80.7237464522} & \num{0.051809138903}  & \numo{92.9793440555} & \num{0.0743589615719}  & \numo{22.6534414443} & \num{0.104814840773}  & \numo{67.1408681261}  & \num{0.324647951038} & \numo{65.8743500195}   \\ 
\malopa & \num{0.0938506078623}  & \numo{80.9504099653} & \num{0.0514183959982}  & \numo{93.2710501419} & \num{0.0743737252308}  & {\bf \numo{22.7531554226}} & \num{0.104432387643}  & \numo{67.8830817345}  & \num{0.324075116734} & {\bf \numo{66.2144243161}}   \\ 
\fenda & \num{0.0919256223791}  & \numo{80.7927309997} & \num{0.0515912809523}  & \numo{93.0838063702} & \num{0.0741486230276}  & \numo{22.7505313705} & \num{0.1054431006}  & \numo{67.7325164222}  & \num{0.323108626959} & \numo{66.0898962907}   \\ 
\bottomrule
\end{tabular}
\caption{Aggregated Multidomain/multitask results of tasks in \cref{tbl:featuredesc}: {\bf bold} indicates best average results over all models. %
}
\label{tbl:mdmt-average}
\end{table*}

We compare our reparametrized models against the following feature-reparametrizing approaches:
\begin{description}
\item[\malopa] For each task $t$, we concatenate the word embeddings $\vec{emb}(w)$ with task embeddings $\vec{k}_t$: $[\vec{emb}(w); \vec{k}_t]$. $\vec{k}_t$ are trained along with the network, and hopefully contains task-relevant information.
This idea originated from the \malopa\,(MAny Language One PArser) parser \cite{ammar2016}.
\item[\fenda]
In this setting, each task has its own predictor and two message encoders, one shared and the other specific to itself. The two encoder outputs are concatenated, linearly transformed, and fed into the predictor. This is an adaptation of the \fenda (Frustratingly Easy Neural Domain Adaption) model in
\citep{kim2016}, which in turn is a neural extension of the classic paper by \citet{daume2007}.
\end{description}
We also compare them against the two baselines:
\begin{description}
\item[\disjoint] Each task has its own predictor, thread encoder, and message encoder.
\item[\tied] Each task has its own predictor. All tasks share the same thread encoder and message encoder. As we noted in \cref{sec:reparametrization} it has been empirically found that the model is capable of learning a shared representation across tasks and  domains.\cite{glorot2011}
\end{description}
We evaluate our proposed models, feature-reparametrizing models, and the non-domain-adaptive baselines on tasks listed in \cref{tbl:featuredesc} in these following multidomain/multitask transfer settings:
{\bf (E)}, {\bf (E+I)}, {\bf (E+R)}, {\bf (I+R)}, {\bf (E+I+R)}, where {\bf E}=Email, {\bf I}=IRC, {\bf R}=Reddit. Note that since only the emails have two meta features E-A and E-T, we have {\bf (E)} as our only multitask transfer setting. The results are in \cref{tbl:mdmt-average}.
Difference between results from all models is small. We inspected the model outputs and found they all suffer severely from the label bias problem --- all four tasks have very unbalanced label distributions; and the network learns to strongly favor the more frequent label. The label bias problem can potentially be addressed by using a globally normalized model which we leave as future work.
Despite the small margins, we can see that both model- and feature-reparametrizing models outperform the baselines in terms of likelihood. Moreover, our reparametrized models consistently achieve higher likelihood than baselines on test data in all transfer settings. In addition, \add\, and \addmul\, perform comparably well against strong domain-adaptive models in terms of accuracy. 

\subsection{Recipient Action with Minimal Supervision}
\label{sec:minsup}
\begin{table}[ht]
\small
\centering
\begin{tabular}{cr}
\toprule
Setting  & $F_1$ \\
\midrule
\add & \numo{32.7241147628}\sym{*} \\
\addmul & \numo{32.5742784297} \\
\affine & \numo{31.6653789348} \\
  \midrule
\disjoint & \numo{27.8694132587} \\
\tied & \numo{30.7020859959} \\
\malopa & \numo{30.6903287165} \\
\fenda & \numo{31.4313323055} \\
\midrule
\doctvec  & \numo{26.7319113957}   \\
\bottomrule
\end{tabular}
\caption{Results of \cref{sec:minsup}. \add\,is significantly outperforming the best baseline \fenda ($p=\num{0.044286491784080818}$) and \addmul\, borderline significant ($p=\num{0.08163768966141785}$) against \fenda\,, the best-performing domain-adaptive baseline model under paired $t$-test. The difference between \add and \addmul\, against other baseline models are also significant under paired $t$-test. Hyperparameters are regularization strength $C$ and transfer setting.}
\label{tbl:two-phase}
\end{table}
\begin{table}[ht]
\small
\centering
\begin{tabular}{crrrr}
\toprule
 & \multicolumn{4}{c}{$F_1$} \\
  \cmidrule(lr){2-5} \
 Setting & E & E+I+R & E+R  & E+I \\
 \midrule
\add & \numo{24.3328051165} & \numo{30.1534183525} & \numo{25.9690487907} & \numo{32.6839251657} \\
\addmul & \numo{22.8737340439} & \numo{30.2506606426}  & \numo{27.8310546096} & \numo{33.0636945739} \\
\affine & \numo{30.8249878622} & \numo{28.3745130702} & \numo{26.554399144} & \numo{33.0876984225} \\
\midrule
\disjoint & \numo{27.6358589692} & \numo{29.2908998428} & \numo{26.2468863835} & \numo{25.7771756349} \\
\tied & \numo{27.2349753625} & \numo{31.2022182115} & \numo{25.2450905502} & \numo{30.8631942751} \\
\bottomrule
\end{tabular}
\caption{Breakdown on different transfer settings.}
\label{tbl:breakdown}
\end{table}
We now turn to a task-based evaluation where we use our extracted thread embeddings on the task of predicting an email recipient's next action. In particular, we focus on scenarios where we do not have a sizable amount of annotated data to train a neural network in an end-to-end fashion, and when we simply did not anticipate the task when we trained the model. 
This setting evaluates the network's ability to generalize
over multiple tasks and learn a good representation.

To be more specific, the setup is as follows: we use trained models from \cref{sec:end-to-end} to encode thread embeddings from action-annotated emails $\{\vec{m}_a\}$ of \cref{sec:data}.
Subsequently we use these thread embeddings to train $L_2$-regularized logistic regression classifiers for the action labels.
We compare them against  classifiers trained with features extracted from the baselines \tied, \disjoint, \malopa, and \fenda. We also compare it against doc2vec embeddings trained on the whole Avocado corpus (listed in \cref{tbl:two-phase} as \doctvec).

Given the small size of annotated data, we decide to evaluate the models with nested cross validation (CV).  In the outer layer, we randomly split the annotated emails into (train+dev)-test splits,\footnote{120 splits with a ratio of $(0.67, 0.33)$} in a thread-wise fashion. In the inner layer, we use 7-fold CV on the (train+dev) split to find the best hyperparameters. The best hyperparameters are then used to train a classifier, which is subsequently evaluated on the test split of the outer layer CV. We report the average $F_1$ in \cref{tbl:two-phase}. 
\disjoint\, performs poorly on this task since there is no baked-in constraint for it to learn a shared representation. All shared-representation baselines (\tied, \fenda, \malopa) performed better than both \disjoint\, and \doctvec. Still, our reparametrized models compare favorably against the feature-reparametrizing baselines. 

We do another cross validation evaluation, over different transfer settings in \cref{tbl:breakdown}. It seems that while both Reddit {\bf (E+R)} and the IRC {\bf (E+I)} datasets do better than email only {\bf (E)}, the IRC dataset is much more helpful than Reddit. This resonates with our initial findings in \cref{sec:outdomain} that the IRC dataset is more similar to emails. 
We note that all the $F_1$ scores are low. Nonetheless we find it encouraging that out-of-domain data is able to help learn a better representation in this extremely resource-scarce setting. 

\section{Related Work}
\label{sec:related-work}
There has been a lot of work on multidomain/task learning with
shared representation as we described in \cref{sec:introduction}.
Our work is also closely related to work on email speech act modeling and recognition \cite{cohen04,lampert2008,jeong2009,defelice2012}.
The idea of model reparametrization for domain adaption is abundant in the literature of hierarchical Bayesian modeling, such as \citet{finkel2009,eisenstein2011}.

Within the deep learning literature, our work is also related to work on DNN reparametrization for multitask learning, such as \citet{spieckermann2014,yang2016b}. Our work shows the reparametrization approach also works for domain adaptation. 
Finally we would like to point out that \citet{ha2016} introduces an alternative and much more sophisticated reparametrization of RNNs.
An interesting future direction of our work is to follow this work by reparametrizing networks as hypernetworks that take a task embedding as an input. In that case, using the terminology introduced in this paper, we will be feature-reparametrizing the hypernetwork; which in turn model-reparametrizes an RNN.

\section{Conclusion}
In this paper, we have introduced an email recipient action annotation scheme, and a dataset annotated according to this scheme. By annotating the recipient action rather than the sender's intent, our taxonomy is agnostic to specific speech act theories, and arguably more suitable for training systems that suggest such actions. We have curated an annotated dataset, which achieved good inter-annotator agreement levels.
We have also introduced a hierarchical threaded message model \rainbow\, to model such emails. To cope the problem of data scarcity, we have introduced RNN reparametrization as an approach to domain adaptation, and applied it onto the problem of email recipient action modeling. It is competitive against common feature-reparametrized neural models when trained in an end-to-end fashion. We also show that while it is not explicitly designed to encode a shared representation across tasks and domains, it learns to generalize in a minimally supervised scenario.
There are many possible future directions of our work. For example, with appropriate software, we can obtain more annotation automatically, and possibly learn the taxonomy along. Also our reparametrization framework is quite extensible. For instance, user-specific parameters for each user can be learned for personalized models, as in \citet{li2016}. %

\bibliography{conversation-embedding}
\bibliographystyle{aaai}
\end{document}